# ILiAD: An Interactive Corpus for Linguistic Annotated Data from Twitter Posts

## Simon Gonzalez

The Australian National University,
Canberra, Australian Capital Territory, Australia
u1037706@anu.edu.au

**Abstract**

Social Media platforms have offered invaluable opportunities for linguistic research. The availability of up-to-date data, coming from any part in the world, and coming from natural contexts, has allowed researchers to study language in real time. One of the fields that has made great use of social media platforms is Corpus Linguistics. There is currently a wide range of projects which have been able to successfully create corpora from social media. In this paper, we present the development and deployment of a linguistic corpus from Twitter posts in English, coming from 26 news agencies and 27 individuals. The main goal was to create a fully annotated English corpus for linguistic analysis. We include information on morphology and syntax, as well as NLP features such as tokenization, lemmas, and n-grams. The information is presented through a range of powerful visualisations for users to explore linguistic patterns in the corpus. With this tool, we aim to contribute to the area of language technologies applied to linguistic research.

## 1. Introduction

In this current age, the use of social media platforms has permeated across all circles of society, from personal communication to government communications. Its impact is hard to be overstated. It is considered as a form of mass media, but distinctive from other forms such as television and radio, where the information is presented by a specific broadcasting mechanism (Page et al., 2014). In the case of social media, the content can be delivered by anyone, making it more personal and individual than other mass forms. The adopting of this technology in language research has been an organic and necessary process. This is because language research investigates the use of language in society, and since social media is a medium of language, we need to understand how we use language in this digital world.

One framework that has efficiently paved the way for linguists to examine social media language is Computer-Mediated Communication (CMC). This has been defined as a relatively new 'genre' of communication (Herring, 2001), which can involve a diversity of forms connecting the physical and the digital (Boyd and Heer, 2006). One of the focus of study in CMC research is on the intrinsic characteristics of digital language, e.g. stylistics, use of words, semantics, and other relevant linguistic features. This has been done for various CMC types, including social media.

But describing social media features is not a straight-forward task because it is not a homogenous genre. It has a diversity of types depending on the main shareable content (e.g., YouTube for videos, Twitter for texts[1]) and main format (e.g., Reddit as a discussion forum, Pinterest Product pins for products purchase), for example. But one common feature is that all platforms have an interactive component in which users can express ideas, comment, and reply to other people's perspectives. The inherent communicative aspect in this social interaction is one that has strong implications in linguistic research, which is that when we analyse language from social media, we look at how language is used in natural contexts, with concrete communicational purposes. What distinguishes then our approach as language researchers, from engineers and app developers, for example, is that we are interested to study how people use technology to communicate and describe what makes it a distinctive type of language (Page et al., 2014). In this sense, we are interested in identifying the language patterns as used in social media platforms, knowing that patterns found in social media are not necessarily representative of language patterns in other contexts. This has been demonstrated empirically by Grieve et al. (2019), where they compared Twitter data versus traditional survey data. They found that some patterns were observed more strongly in Twitter data than in the survey data. Results like these are evidence then that when we deal with social media language, we are examining a way of expression, which has features like other language forms, but at the same time it has its own distinctive characteristics. This is paramount to be considered when new language analysis technologies are developed.

### 1.1. Twitter and Corpus Linguistics

The combination of language research and social media is a complex endeavor, making people working with both apply skills that are necessary in this interdisciplinary undertaking. One area that reflects this complexity and that has efficiently adapted social media is Corpus Linguistics (CL). A strong characteristic of CL is that it is used to collect, store, and facilitate language analysis for large datasets (Szmrecsanyi, 2011; Grieve, 2015). And with the advantage of having more sophisticated tools available, such as in social media research, corpora are becoming larger and larger, with the only constraints being computational power and storage capacity.

Many social media platforms have been widely used for language and linguistic research (c.f. Liew and Hassan, 2021; Nagase et al., 2021; Trius and Papka, 2022; Wincana et al, 2022). Out of these platforms, Twitter stands out due to its world spread, and the option it gives to researchers when stratifying the demographics of user accounts, including the use of the geo-code and time-stamp information of the posts[2] (Grieve et al., 2018). It is classified as a microblogging site (Chandra et al. 2021) where the content can be on opinions, news, arguments, and other types of sentences (Chaturvedi et al., 2018). Because of their wide-spread use, it has been used in the creation of numerous corpora created from Twitter posts (c.f. Dijkstra et al., 2021; Grieve et al., 2019, Tellez et al., 2021).

---

[1] The type of content of social media platforms is not restricted to only one. This is just an example on the main purpose for specific cases. For instance, YouTube allows users to write comments on videos and Twitter can embed videos on posts.

### 1.2. Current Project

In this paper, we present the development of a web-based corpus from Twitter posts, named *ILiAD: An Interactive Corpus for Linguistic Annotated Data*. In relation to our methodological approach, we propose that corpora built from social media helps study the patterns of language used in this context and capture their linguistic complexity. By doing this, we can have a better view of the multilayered nature of the corpus.

## 2. Goal of the Paper

The aim of the corpus is to capture the linguistic complexities used in Twitter language, and we chose two types of account users: news agencies and individuals. We explore the differences between their structures and patterns. The language of journalism is characterised based on its main purpose: exert influence on readers and convince them on a specific interpretation (Fer, 2018; Moschonas, 2014). This is achieved by three main stylistic features. The first one is language clarity, a feature that is strongly appropriate for journalism more than many other language styles. The second one is accuracy. This refers to the ability to convey ideas accurately and avoiding ambiguities in interpretation. The final one is the simplicity. This aims to convey messages without the use of complex words that may blur the intention of the message (Fer, 2018). The aim therefore is to prepare the corpus for further exploration, querying and analysis to understand the language used in Twitter.

The analysis can focus on many linguistic parameters and here we approach it in an integrated way. This can give users the opportunity to explore the corpus from different angles and linguistic perspectives.

## 3. Methodology

The stages of data collection, data processing, and app deployment were carried out in R (R Core Team, 2021), using *shiny* R (Chang et al., 2021) for the app development. Apps developed in shiny have three main advantages. The first one is its interactivity capability. With this, users can interact with the whole corpus, across a range of visualisation outputs and tables. The second one is its reactive power. With this, users modify parameters in the tables and visualisations, and the app changes outputs based on user inputs. The positive impact on corpus linguistics is invaluable. With these features, a corpus can be used to have a full understanding on the shape of the data as well as an exploration of patterns.

### 3.1. Data Collection

We applied four criteria to identify the Twitter accounts to be included in the corpus. The first criterion was that account users (news agencies and individuals) had to have English as the main language of communication. The second one was that accounts had to be active at the moment of extraction. The reason was to capture tweets that were synchronous and where topics and trends could be shared across accounts. The third criterion was that accounts had to have a large number of tweets, enough to reach over 3,000. This was done to make sure that enough posts were left after the filters were applied, which is explained below. The final criterion was to only include those users whose posts were not mainly retweets. This filter aimed to exclude those accounts that do not produce content but only retweet posts from other accounts. From this, we identified 29 news agencies, and 27 individual accounts. The percentages are shown in Table 1.

| User Type | Total Tweets | Percentage |
|---|---|---|
| *News Agency* | 84,354 | 54% |
| *Individual* | 71,477 | 46% |
| **Total** | **155,831** | **100%** |

Table 1: Total number of tweets in the corpus and their proportions per account type.

The data extraction was done through an R script developed by the main author. We used the *rTweet* (Kearney, 2019) package, which allows users to gather Twitter posts by the free Twitter API, giving a total of over 156,000 tweets.

| Year | Total Tweets | Percentage |
|---|---|---|
| *2009* | 139 | 0.1% |
| *2010* | 178 | 0.1% |
| *2011* | 497 | 0.3% |
| *2012* | 2230 | 1.4% |
| *2013* | 5097 | 3.3% |
| *2014* | 3625 | 2.3% |
| *2015* | 5159 | 3.3% |
| *2016* | 6745 | 4.3% |
| *2017* | 5508 | 3.5% |
| *2018* | 6301 | 4.0% |
| *2019* | 7847 | 5.0% |
| *2020* | 18742 | 12.0% |
| *2021* | 20697 | 13.3% |
| *2022* | 73066 | 46.9% |
| **Total** | **155,831** | **100%** |

Table 2: Total number of tweets in the corpus and their proportions per year.

### 3.2. Data Processing

From the collected data, we applied six filters to make sure that the corpus reflects comparable linguistic data for all account users. The first filter was to exclude tweets that were not in English (n=10,067; 6%). This was done by filtering out those tweets which did not have the English (*en*) assigned by Twitter's machine language detection, which is annotated in the tweet's metadata. The second filter was to exclude re-tweets (n=23260; 15%). This restricts the data only to those posts that come from the given user and not from other accounts. The third filter was to exclude quote tweets (n=7,142; 5%). These are tweets that are re-tweeted with an added comment from the user. Keeping quote tweets in the data would add repeated tweets to the corpus and also would add patterns and word counts that do not correspond to a specified account. The fourth filter deleted repeated tweets (n=778; 0.5%). This targeted those cases in which account users write the same content and post it as a separate tweet, but not as a re-tweet. Similar to quote tweets, keeping repeated tweets would inflate the content of the corpus and it would not be representative. For the fifth filter, we excluded strings that were URL links, which do not have linguistic features[3] of interest in this paper (n=1,208; 0.8%). For the sixth and last filter, we first calculated the number of words for each tweet, which were split by white spaces to get the number of individual words. We then excluded those tweets that had a length of less than eight words (n=14,125; 9%). This filter targets those tweets which do not have linguistic content but only social media features such as hashtags or links.

---

[2] The geo-code information is optional in Twitter, and the user decides whether to show this or not. Other approaches include running algorithms that identify locations based on factors such as time zone and language features, which are used to infer locations.

With these filters, the final data contained 112,690 tweets. This is a loss of 28% (n = 43,919) of the original data exported from the Twitter API.

## 3.3. Text Processing

After data filtering, we implemented a wide range of *Natural Language Processing* (NLP) techniques for the data wrangling and analysis. We carried out the text processing using the *UDPipe* (Straka and Straková, 2017) package as the main tool for the NLP tasks. UDPipe is defined as single tool which contains a tokenizer, morphological analyzer, Parts-Of-Speech tagger, lemmatizer, and a dependency parser. It currently offers 77 language models, with some languages having more than one model available. We used the *EWT* English model available in the package. We selected the text column from the API output and made it the input for the main UDPipe function. The core purpose of the UDPipe package is to create a single-model tool for a given language which can be used to process raw text and convert it to a CoNLL-U-formatted text. This format stores tagged information for all words in dependency trees, including morphological and syntactic features (Straka and Straková, 2017). From this format, the UDPipe algorithm creates morphological taggers and dependency parsers. The main taggers are described below.

### 3.3.1. Tokenization

The tokenization tools are wrapped within a trainable tokenizer based on artificial neural networks, specifically, the bidirectional LSTM artificial neural network (Graves and Schmidhuber, 2005) and a gated linear unit – GRU (Cho et al., 2014). It works by comparing the words in the input text to the trained tokenizer and does not add any additional knowledge about the language. If a given word, or group of words, is not recognized, the tokenizer tries to reconstruct it by utilizing an additional raw text corpus.

### 3.3.2. Morphological Analysis

There are three main fields tagged in the data process:

1. Part-of-speech tagging
2. Morphological features
3. Lemma or stem

The parts-of-speech tagging uses *MorphoDiTa* (Straková et al., 2014). The tagging process exploits the rich linguistic features of inflective languages with large number of suffixes, where multiple forms can be related to a single lemma. From this, the tagger estimates common patterns on endings and creates morphological templates from the observed clusters. On Table 3, we show the top ten counts and proportions of Parts-Of-Speech tags in the current corpus, as output from UDPipe.

| POS | Corpus Count | Percentage |
|---|---|---|
| *NOUN* | 76,795 | 20.8% |
| *VERB* | 62,537 | 16.9% |
| *ADP* | 39,237 | 10.6% |
| *PROPN* | 37,862 | 10.3% |
| *PRON* | 37,399 | 10.1% |
| *DET* | 31,284 | 8.5% |
| *PUNCT* | 30,001 | 8.1% |
| *ADJ* | 24,452 | 6.6% |
| *ADV* | 16,425 | 4.4% |
| *AUX* | 13,171 | 3.6% |

Table 3: Total number of top ten Part-Of-Speech tags in the corpus and their proportions.

### 3.3.3. Classification Features

UDPipe uses two models that facilitate the tagging process and improve the overall accuracy by employing different classification feature sets. The first one the POS tagger, which disambiguates all available morphological fields in the data. The second model, a lemmatizer, disambiguates the lemmas tagged.

### 3.3.4. Dependency Parsing

Dependency parsers are part of the family of grammar formalisms called *dependency grammars* (Jurafsky and Martin, 2021). In these, the syntactic structure sentences are described on the grammatical relations between the words, shown as directed binary dependencies. All structures start at the root node of the tree, and then components and the dependencies are shown throughout the entire structure. Dependency parser trees can deal very efficiently with languages that are rich morphologically and also have a relatively free word order, for example Spanish, Czech, and English, with varying flexibility. Another important advantage of using dependency parsers is that they allow closer examination of semantic relationships between arguments in the sentence.

Summing up, the features, descriptions, and tagging done by the UDPipe framework, offer invaluable information relevant for linguistic analysis used in Corpus Linguistics. With these features extracted for all tweets, we have information available at different layers for linguistic

---

[3] URL Links are an important aspect of social media language. However, its analysis is beyond the scope of this paper.

analysis: morphological, syntactic, and even semantic, through the dependency parsers.

### 3.4. Data Filtering

After obtaining the output from the UDPipe package, we proceeded to filter the data. The motivation was to prepare it for the linguistic analysis within the corpus. This filtering process affects two dataset outputs which used for different purposes in the corpus. The first one is used for calculating n-grams and word frequencies. The second one is for showing Syntactic Dependencies.

#### 3.4.1. Token Filtering

Identifying the right tokens in social media language is a difficult process. The correct practice in this step is crucial to achieve efficient outcomes. This filtering differs from the practice done on other language media such as the language in newspapers, television, and academic papers. Following O'Connor et al., (2010), we excluded tokens containing hashtags, URL links, @-replies, strings of punctuation, and emoticons[4]. Their proportions are shown in Table 4.

| Content Excluded | Total Count | Percentage |
| --- | --- | --- |
| Emoticons | 1,556 | 0.4% |
| Hashtags | 1,986 | 0.5% |
| URL Links | 2,857 | 0.7% |
| @-replies | 3,851 | 0.9% |
| Punctuation | 30,001 | 7.3% |

Table 4: Total number of social media content excluded and their proportions in the whole corpus.

#### 3.4.2. Removing Stop Words

Following standard procedures, we removed stop words for calculating n-grams and word frequencies. An important observation is that removing stop words is a compromise for the corpus, since certain word combinations are affected, especially those which appear together with the words in the list. Future versions of this work aim to efficiently implement analysis considering the role of stop words in the corpus.

Here we removed stop words by following the steps below:

1. First, we selected a list of stops words from the *stopwords* (Benoit et al., 2021) package in R. We selected the ones used for English and it included 175 words (see Table 5 for the top 15).
2. Next, we filtered out the stop words in this data subset.
3. Finally, we filtered out stop words that are specific for Twitter, and that includes words such as *RT*, *follow*, *follows*, and *following*. In future versions, we aim to implement a disambiguation algorithm where a key word, such as follow, can be identified as a word used in social media context (e.g. *follow us on Twitter*), or in a more traditional one (e.g. *follow the road*).

| Stop word | Total Count | Percentage |
| --- | --- | --- |
| *the* | 17,151 | 4.18% |
| *to* | 11,543 | 2.81% |
| *a* | 9,076 | 2.21% |
| *be* | 8,480 | 2.07% |
| *and* | 7,844 | 1.91% |
| *of* | 7,001 | 1.71% |
| *I* | 6,734 | 1.64% |
| *in* | 6,429 | 1.57% |
| *you* | 5,315 | 1.3% |
| *have* | 4,083 | 1% |
| *that* | 3,933 | 0.96% |
| *it* | 3,803 | 0.93% |
| *for* | 3,793 | 0.92% |
| *on* | 3,552 | 0.87% |
| *he* | 3,442 | 0.84% |

Table 5: Top 15 stop words excluded in the data subset and their proportions in the corpus.

#### 3.4.3. Sentence Structure Filtering

In this filter, we aimed to identify those posts which were not linguistic phrases or sentences, thus including only those structures that were classified into a sentence category. For each of the tweet breakdown done by UDPipe (as shown in Table 6), we looked at the PUNCT classification, where we identified three types of sentences: statements (ending with "."), questions (ending with "?") and exclamations (ending with "!"). Any unclassified sentence was deleted from the dataset. Deciding to keep sentences that follow the standard punctuation symbols has a strong impact in a corpus based on Twitter language, since sentences here can follow other rules, e.g. ending a sentence with emoticons or other use of punctuation symbols, such as !!! or :). However, an important number of sentences follow the most standard use of punctuation symbols, which is a reliable representation of the data collected. Finally, for each sentence, we checked whether there was a conjugated verb. For those sentences which had no conjugated verbs, the identified sentence was deleted from the dataset used for the Syntactic Dependency section. For this, we created a data subset that only contained sentences and their corresponding classification done in the previous step. This was the input for the Section explained in 4.1.2.

| token | upos | feats | dep_rel |
| --- | --- | --- | --- |
| *Senate* | PROPN | Number=Sing | nsubj |
| *needs* | VERB | Mood=Ind | root |
| *to* | PART | | mark |
| *think* | VERB | VerbForm=Inf | xcomp |
| *and* | CCONJ | | cc |
| *vote* | VERB | VerbForm=Inf | conj |

Table 6: Sample output from *UDPipe*.

### 3.5. Calculating N-Grams

By implementing NLP techniques, this brings more depth to the corpora analysis since it allows users to explore more areas in the data. In the current version of the app, we use unigram and bigram explorations. The n-grams are calculated using the *tidytext* (Silge and Robinson, 2016) package. We followed the established approach of deleting stops words in English, using the *stopwords* package. After the filtering, the n-grams were calculated across all the data.

---

[4] Emoticons entail rich linguistic information. However, their analysis is not included in this version of the tool.

### 3.6. Entity Identification

A second group of NLP techniques implemented is the identification of entities in the corpus, and that includes mentions of people, physical locations, and established organisations. We used the *entity* (Rinker, 2017) package for this purpose. This package is a wrapper to simplify and extend the *NLP* (Hornik, 2020) package and the *openNLP* (Hornik, 2019) package named entity recognition. The advantage of this approach is that we can use it to detect important information, which is crucial especially in large datasets, that can be captured when identifying entities. This also has a strong impact on our understanding of linguistic features, since they are related to important elements in sentences, such as nouns and adjectives. By implementing this, the app brings more depth to the corpora analysis since it allows users to explore the main entities in the corpus.

### 3.7. Twitter Metrics

The final metrics measured and obtained aims to show information that is relevant when dealing with Twitter data. The motivation is to be able to contextualise the information in the corpus within the overall world of social media. The information presented here is extracted from the Twitter API output, which means that we display two features publicly available. The first one is the number of tweets across time. We also include a general summary of the main sour locations by country of the tweets contributing to the data. Previous studies (c.f. Grieve et al., 2019) have demonstrated that the use of geo-coding information is relevant for linguistic studies, but here, we only show the country of origin of all tweets without identifying individuals or linking linguistic features to any demographics.

## 4. App Infrastructure

The app was developed in RStudio, which has been widely used for corpus linguistics development and related tasks (Abeille and Godard, 2000; Gries, 2009), and the main framework was within shiny R. Shiny apps allow great interactivity and responsiveness. Interactivity allows users to explore visualizations in effective ways, and responsiveness allows users to navigate contents in real time, with the use of clicks and dropdown menus. Other libraries that we used for the creation of visuals were *ggplot2* (Wickham, 2016) and *echarts4r* (Coene, 2022). ggplot2 allows a great degree of flexibility when creating figures. This is relevant since there is a lot of complexity of the linguistic data that we present. But this allows complex ideas to be presented in a digestible way. Another advantage of this is that it allows users to see data points within the general context as well as being able to narrow down into more specific analysis. This creates a seamless navigation of linguistic data in an efficient way. The presentation of the app components was divided into two main sections. The first component gives users tools to explore linguistic features and the second one gives information on Twitter metrics. Due to the limitations on Twitter Terms of Service, the app cannot display the raw tweets as a database format nor give the option to download data. The interactive tool therefore focuses on the presentation of the linguistic features derived from the data.

### 4.1. Exploring Calculated Features

The linguistic features are the main backbone of the corpus. In this section, there are visualisation options that can be used to have both a broad understanding of patterns, as well as a deep exploration of linguistic features.

#### 4.1.1. Parts of Speech

This section gives the overall statistics of the words classified into their POS, including distributions and proportions per year and sentence type. The exploration can be done in different levels: all corpus or by user type (news agencies or individuals). The input data in this section comes from the **Sentence Structure Filtering Section (3.4.3)**.

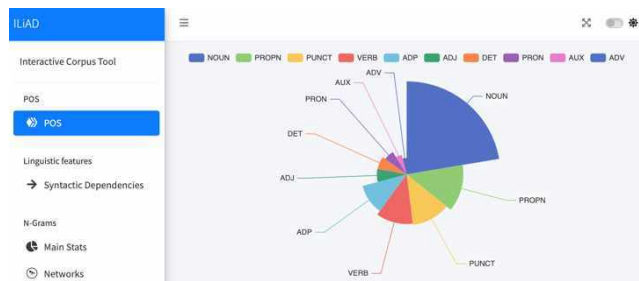

Figure 1: POS Distributions Tab.

#### 4.1.2. Syntactic Dependencies

This section allows users to explore the syntactic dependencies of all the available sentences. Here we use a combination of the UDPipe output and the *textplot* (Wijffels et al., 2021) package, which creates the dependencies as in the figure below. Since users can select all available sentences, this is a powerful function than can be used to explore syntactic patterns across the corpus and facilitates the understanding of syntactic structures. The input data in this section comes from the **Sentence Structure Filtering Section (3.4.3).**

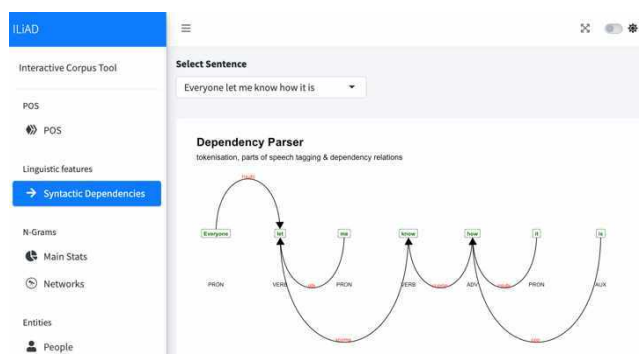

Figure 2: Syntactic Dependencies Tab.

#### 4.1.3. Exploring N-Grams

N-grams are explored through visualisations, including connection networks. These networks are developed within the *Network Analysis* (NA) approach. The power of this analysis comes from its capability of observing relationships between components. This technique has been implemented in other fields, such as psychology (Jones et al., 2021; Mullarkey at al., 2019), and social network research (Clifton and Webster, 2017; Würschinger, 2021). NA visualizations follow the assumption that if a relationship is meaningful within the whole network, it will stand out from other relationships by stronger connections than random or weaker relationships. In this analysis, the connections are based on the frequencies which connect n-grams. Here we use the functionality from the *visNetwork* (Almende, 2021) package.

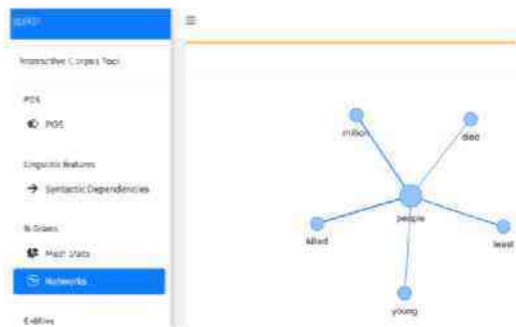

Figure 3: N-Grams Visualisation Tab showing a network relationships.

#### 4.1.4. Exploring Entities

We use a different visualization approach for the entities captured in the corpus. We use bar plots and word clouds. The advantage of bar plots is that they show the frequencies in a way that we can see from the most frequent to the least frequent, organized from left (most frequent) to right (least frequent). Word clouds are an easy and user-friendly way to represent frequencies. Here, more frequent words are represented with larger fonts than less frequent words. An example for the organizations mentioned in the corpus is shown in the figure below. At the top, we see the bar plot and at the bottom the word cloud.

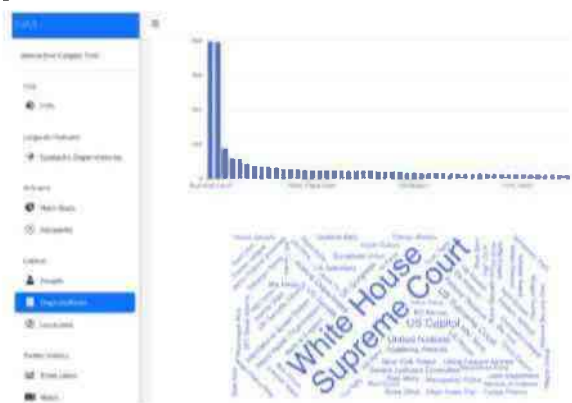

Figure 4: Named Entities Visualisation Tab.

### 4.2. Twitter Data Metrics

The final section shows relevant Twitter data metrics, for which we dedicate two sections. The first one is a timeline visualization using a combination of the *ggplot2* package and the *plotly* (Sievert, 2020) package. This combination gives ggplot2 plots interactive power. The timeline displays the number of posts across time, for all the data available in the corpus. This timeline can also be selected to observe by account type, giving more granularity of exploration. Another timeline visualization is applied to N-grams. This has been used to observe lexical innovations (c.f. Grieve et al., 2019), by looking at N-grams that increase in terms of frequency across time. This tool can facilitate this type of analysis.

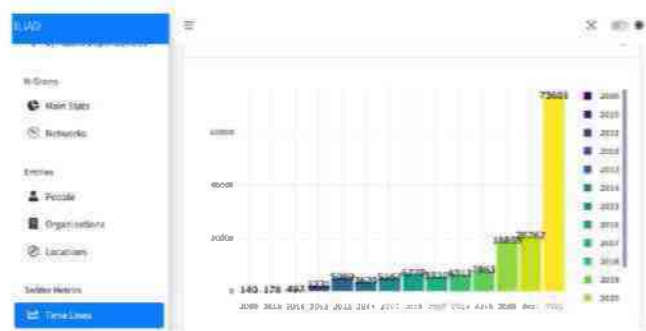

Figure 5: Twitter Timeseries Count Tab.

The second visualization implemented is a world map showing the region source information of tweets. The purpose is to visualize the main geographical areas from where the tweets come. We use the functionality from *echarts4r* package, which is very efficient at displaying this type of information, as well as being interactive in an online context.

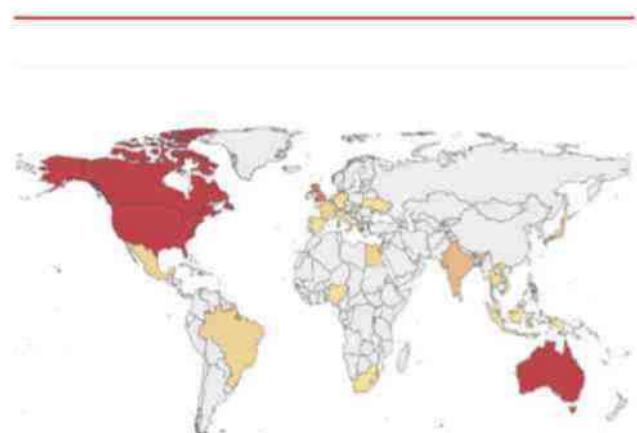

Figure 6: Twitter Map Tab.

### 5. Discussion

The app presents a wide range of visualizations and analyses from the Twitter corpus. The features capture different linguistic layers, including morphology, syntax, and n-grams. With the inclusion of Twitter metrics, this tool gives all exploration opportunities to understand the whole corpus. R and shiny R have proven to be an efficient combination to develop and deploy the corpus. For the text processing tasks, the use of the UDPipe and tidytext packages have been highly effective. The in-built functions have been used and we have created our custom-made functions to complete the tasks done throughout the whole process. For visualization tasks, the combination of ggplot2, plotly, visNetwork, and echarts4r has demonstrated efficient to represent complex linguistic features and relationship analysis. The app can be accessed through the following GitHub repository: https://github.com/simongonzalez/ILiAD.

## 6. Conclusion

In this paper, we have presented the development of a linguistic corpus based on the Twitter posts. It has been designed to be used by a diversity of audiences who are interested in exploring linguistic patterns from corpora based on social media language. Similar tools have been developed with invaluable contributions to the field of Corpus Linguistics. Our proposal, however, makes stronger integrations with a variety of visualization types that enhance the analysis in a holistic way. The tool also gives users interactive and reactive power throughout all the data, which not only offers a corpus to analyse, but a corpus to interact with and query in a more organic way, compared to more traditional approaches of presenting corpora. Finally, it has been developed within an open-source framework, making it freely available to any user interested in using and even expanding this tool.

## 7. Future Work

In the current version, we have selected a relatively small number of users in the corpus, as compared to other larger projects with similar goals. This is to allow the implementation of the interactive capability in the visualization methods, which requires a high level of computational power. We aim to add more data in future versions using more efficient processing algorithms. Finally, we see the value of adding linguistic analysis to emoticons. In a future version, we aim to include analysis on emoticons, as a distinctive component of social media language.

## 8. Acknowledgements

I want to thank the anonymous reviewers of this paper for their invaluable comments and insights in the shape and content of the final version. Their generosity and expertise have improved this paper in innumerable ways and saved me from many errors. Those that inevitably remain are entirely my own responsibility.